\pdfoutput=1
\documentclass[11pt]{article}

\usepackage[]{ACL2023}
\usepackage{times}
\usepackage{latexsym}
\usepackage[T1]{fontenc}
\usepackage[utf8]{inputenc}
\usepackage{microtype}
\usepackage{inconsolata}
\usepackage{booktabs,multirow}
\usepackage{adjustbox}
\usepackage{subfig}
\usepackage{hyperref}
\usepackage{tcolorbox}
\usepackage{xcolor}
\usepackage{amsmath}

\newtheorem{theorem}{Question}

\newtcbox{\blue}[1][]{on line,boxsep=2pt,left=0pt,right=0pt,top=0pt,bottom=0pt,colframe=white,colback=blue!30!white,#1}
\newtcbox{\red}[1][]{on line,boxsep=2pt,left=0pt,right=0pt,top=0pt,bottom=0pt,colframe=white,colback=red!30!white,#1}
\newtcbox{\yellow}[1][]{on line,boxsep=2pt,left=0pt,right=0pt,top=0pt,bottom=0pt,colframe=white,colback=yellow!30!white,#1}
\newtcbox{\green}[1][]{on line,boxsep=2pt,left=0pt,right=0pt,top=0pt,bottom=0pt,colframe=white,colback=green!30!white,#1}
\newtcbox{\orange}[1][]{on line,boxsep=2pt,left=0pt,right=0pt,top=0pt,bottom=0pt,colframe=white,colback=orange!30!white,#1}
\title{What does the Failure to Reason with ``Respectively'' in Zero/Few-Shot Settings Tell Us about Language Models?}

\newcommand{\ku}{$^1$}
\newcommand{\tud}{$^2$}

\author{
Ruixiang Cui\ku, Seolhwa Lee\tud, Daniel Hershcovich\ku, Anders Søgaard\ku \\
{\ku}Department of Computer Science, University of Copenhagen\\  
{\tud}Department of Computer Science, Technical University of Darmstadt \\
\texttt{{\{rc,dh,soegaard\}@di.ku.dk, seolhwa.lee@tu-darmstadt.de }}
}

\begin{document}
\maketitle
\begin{abstract}
Humans can effortlessly understand the coordinate structure of sentences such as ``Niels Bohr and Kurt Cobain were born in Copenhagen and Seattle, \textit{respectively}''. In the context of natural language inference (NLI), we examine how language models (LMs) reason with respective readings \citep{Gawron2004TheSO} from two perspectives: syntactic-semantic and commonsense-world knowledge. We propose a controlled synthetic dataset WikiResNLI and a naturally occurring dataset NatResNLI to encompass various explicit and implicit realizations of ``respectively''. We show that fine-tuned NLI models struggle with understanding such readings without explicit supervision. While few-shot learning is easy in the presence of explicit cues, longer training is required when the reading is evoked implicitly, leaving models to rely on common sense inferences. Furthermore, our fine-grained analysis indicates models fail to generalize across different constructions. To conclude, we demonstrate that LMs still lag behind humans in generalizing to the long tail of linguistic constructions.
\end{abstract}

\section{Introduction}
Transformer-based language models (LMs) \citep{devlin-etal-2019-bert,Raffel2019ExploringTL,Brown2020LanguageMA} induce useful representations for a wide range of natural language understanding (NLU) tasks, including natural language inference \citep[NLI;][]{wang-etal-2018-glue,Hu2020XTREMEAM}, especially in in zero-shot or few-shot settings. To what extent this usefulness results from memorization, generalization or the ability of LMs to draw common sense inferences remains an open question. 

\begin{figure}
    \small
    \fontdimen2\font=1pt
    \fontdimen3\font=0pt
    \fontdimen4\font=0pt
    
    \blue{Bohr} and \red{Cobain} were born in \green{Copenhagen} and \orange{Seattle} \yellow{respectively}.
    
    \vspace{2mm}
    \hrule
    \vspace{2mm}
    
    \blue{Bohr} and \red{Cobain} were born in \green{Copenhagen} and \orange{Seattle}.
    
    \vspace{2mm}
    \hrule
    \vspace{2mm}
    
    \blue{Bohr} was born in \green{Copenhagen}. \\
    \red{Cobain} was born in \orange{Seattle}.
    
    \caption{An example of explicit (top, evoked by ``respectively'') and implicit (middle, with no overt marker) respecitve readings. Humans can infer that both sentences have the same ``cross-serial'' meaning (bottom) by relying on commonsense knowledge (that a person is only born in one location) and world knowledge (that Copenhagen and Seattle are mutually exclusive).}
    \label{fig:example}
\end{figure}

To approach it, the linguistic phenomenon of respective readings \citep{Gawron2004TheSO} serves as an excellent probe. This phenomenon has so far been underexplored in NLP, even though it has been studied extensively in linguistic semantics \citep{mccawley1968d,pullum1982natural,dalrymple1995constraints,eggert2000grammaticality}. In English, ``respectively'' is a rare word\footnote{In terms of frequency, in the British National Corpus, ``respectively'' is ranked 13,606th among 18,089 words, and 233rd among 429 adverbs \cite{leech2014word}.} used to establish a one-to-one mapping between two sets of participants and to distribute predicates over sets \citep{okada1999function}. For example, in Figure~\ref{fig:example}, the first conjunct in the subject corresponds to the first conjunct in the object and the second conjunct in the subject corresponds to the second conjunct in the object. The respective relation is bijective and respects the 
relative order of the elements of two different coordinate expressions; it is, in other words, cross-serial. ``Respectively'' can have different syntactic or semantic properties depending on the context, e.g., as a conjunction or adverb.

In this paper, we investigate how LMs reason with respective readings. We propose two datasets,
WikiResNLI (a controlled synthetic dataset) and NatResNLI (a naturally occurring dataset) to cover various explicit and implicit realizations of ``respectively''. Our research questions are:
\begin{enumerate}
  \item Can NLI models reason with ``respectively'' constructions in zero-shot settings?
  \item Can LMs generalize from explicit to implicit respective readings?
  \item Can LMs generalize from synthetic to natural respective readings?
  \item What cues do LMs leverage for prediction?
\end{enumerate}
We experiment with state-of-the-art LMs and analyze the results to gain insights into the limitations of current models and potential directions for future research. We show that LMs are able to generalize effectively in a few-shot learning scenario when the word ``respectively'' is present. However, when the reading is evoked implicitly, a greater number of training instances are necessary. LMs require significantly more instances to generalize to naturally occurring datasets than humans. In conclusion, our study demonstrates that LMs continue to exhibit a deficit in generalizability to infrequent linguistic constructions with limited coverage in their training data.

\section{Respective Readings}

Respective readings are closely related to several types of readings instantiated by plurals and mass terms: distributive readings, collective readings and cumulative readings \citep{Champollion2015}.

\paragraph{Distributive readings.} These usually refer to the application of a predicate to the subsets of a set or group. As for sentence 1(a), it is equivalent to ``John smiled and Mary smiled''. The reading is available because of the nature of the predicate is \textit{atomic} \citep{winter2002flexibility}, similar instances including ``sing'' and ``sleep''. Distributive reading can be enforced with overt distributive markers, i.e., ``every'' and ``each'' \citep{scha1984distributive}. In example 1(b), we enforce the reading by adding ``each'' at the end of the sentence so as to rule out the reading ``John and Mary earn 200 dollars together''.

\begin{enumerate}
\item
\begin{enumerate}
  \item \small{\textbf{Distributive reading:} John and Mary smiled.}
  \item \small{\textbf{Distributive reading with an enforced marker:} John and Mary earn 200 dollars \textit{each}.}
\end{enumerate}
\end{enumerate}

\paragraph{Collective readings.} These are the opposite of distributive readings in that the predicates apply to the whole plural entity instead of individuals. The quantifiers ``all'' and ``most'' instead of ``every'' and ``each'' are usually compatible with collective readings as in example 2(b) \citep{dowty1987collective}.

\begin{enumerate}
\setcounter{enumi}{1}
\item
\begin{enumerate}
  \item \small{\textbf{Collective reading:} The men gathered.}
  \item \textbf{Collective reading with overt marker:} \textit{All} of the men gathered.
\end{enumerate}
\end{enumerate}

\paragraph{Cumulative readings.} These involves two entities but in a symmetric non-scopal relation as in the canonical example 3 \citep{scha1984distributive}. The sentence can be paraphrased into ``There are three boys and two girls, each of the three boy saw at least one of the two girls, and each of the two girls was seen by at least one of the three boys.''. It is discussed sometimes with weak reciprocity \citep{langendoen1978logic}. 

\begin{enumerate}
\setcounter{enumi}{2}
\item\small{\textbf{Cumulative reading:} Three boys saw two girls.}
\end{enumerate}

\paragraph{Respective readings.} These are thought to be a special case of cumulative readings in which a bijective relation holds between the two (or more) sets of entities that enter into the cumulative relation \citep{Chaves2012ConjunctionCA}. 
For example 4(a), the pair (Emiliano Zapata, Morelos) and the pair (Gerhart Münch, Michoacán) are grouped under the \textit{died in} relation.
Respective reading can also arise without the adverb \textit{respectively}, and the absence is even sometimes preferred. As in example 4(b), the binomial expression ``husband and wife'' is so strong that the adverb ``respectively'' is unwarranted.

\begin{enumerate}
\setcounter{enumi}{3}
\item
\begin{enumerate}
  \item \small{\textbf{Respective reading with overt marker:} Emiliano Zapata and Gerhart Münch.} 
died in Morelos and Michoacán, \textit{respectively}.
  \item  \textbf{Respective reading without overt marker:} John and Mary are husband and wife.
\end{enumerate}
\end{enumerate}




\section{An NLI Benchmark for ``Respectively''}
Understanding the coordinate structures in respective readings is effortless for humans, but it remains a question whether LMs, after being pre-trained on billions of tokens and fined-tuned on thousands of NLI instances, can reliably process them. 

To probe LMs' behaviour in the presence of respective readings, we construct two English NLI datasets: WikiResNLI, a synthetic dataset based on an analogy corpus, and NatResNLI, a dataset sourced and created from natural occurrences. We release both datasets on Github\footnote{\url{https://github.com/ruixiangcui/WikiResNLI_NatResNLI}} and describe the detailed creation steps below. 

    
    
    

\begin{table*}[t]
\footnotesize

\centering 
\small 
\begin{tabular}{@{}p{3.2cm}p{2.9cm}p{9.1cm}@{}}

\toprule
& \textbf{Denotation} & \textbf{Natural Language Example} \\
\midrule
\textbf{Premise:} & \blue{$w_1$} and \red{$w_3$} $p$ \green{$w_2$} and \orange{$w_4$}, respectively. & \textit{\blue{Emiliano Zapata} and \red{Gerhart Münch} died in \green{Morelos} and \orange{Michoacán}, respectively} \\
\midrule
\textbf{Hypotheses:} \\
\textbf{Entailment (1)}, 1S1O & \blue{$w_1$} $p$ \green{$w_2$}. & \textit{\blue{Emiliano Zapata} died in \green{Morelos}.} \\
\textbf{Entailment (2)}, 1S1O & \red{$w_3$} $p$ \orange{$w_4$}. & \textit{\red{Gerhart Münch} died in \orange{Michoacán}.} \\
\textbf{Contradiction (1)}, 1S1O & \blue{$w_1$} $p$ \orange{$w_4$}. & \textit{\blue{Emiliano Zapata} died in \orange{Michoacán}.} \\
\textbf{Contradiction (2)}, 1S1O & \red{$w_3$} $p$ \green{$w_2$}. & \textit{\red{Gerhart Münch} died in \green{Morelos}.} \\
\textbf{Contradiction (3)}, 1S2O & \blue{$w_1$} $p$ \green{$w_2$} and \orange{$w_4$}. & \textit{\blue{Emiliano Zapata} died in \green{Morelos} and \orange{Michoacán}.} \\
\textbf{Contradiction (4)}, 1S2O & \red{$w_3$} $p$ \green{$w_2$} and \orange{$w_4$}. & \textit{\red{Gerhart Münch} died in \green{Morelos} and \orange{Michoacán}.} \\
\textbf{Contradiction (5)}, 2S1O & \blue{$w_1$} and \red{$w_3$} $p$ \green{$w_2$}. & \textit{\blue{Emiliano Zapata} and \red{Gerhart Münch} died in \green{Morelos}.} \\
\textbf{Contradiction (6)}, 2S1O & \blue{$w_1$} and \red{$w_3$} $p$ \orange{$w_4$}. & \textit{\blue{Emiliano Zapata} and \red{Gerhart Münch} died in \orange{Michoacán}.} \\

\bottomrule
\end{tabular}
\caption{Example analogy in the spirit of \citet{Garneau2021AnalogyTM}. Both entity pairs \textit{(\blue{$w_1$}, \green{$w_2$}}; \red{$w_3$}, \orange{$w_4$}) share the $p$ relation. Object entities are \textit{unique} in that given an entity pair and a subject, the fourth is uniquely determined. We generate eight hypotheses for each premise: 1S1O refers to one subject and one object, 1S2O refers to one subject and two objects and 2S1O refers to two subjects and one object.}
\label{tab:wikiresnli_example}
\end{table*}

\subsection{Synthetic Dataset: WikiResNLI}
To generate a controlled synthetic challenge set for reasoning with respective readings, we exploit a useful relationship between coordination constructions and \textit{analogies}.
Analogy is concerned with similarities between observable properties and causal similarities. 

\paragraph{Analogy dataset.} \citet{Garneau2021AnalogyTM} proposed
WiQueen, a multilingual analogy dataset consisting of 78,000 analogies extracted from Wikidata. A subset of 9,000 instances is annotated where all four entities are \textit{unique}. These are the analogies in which all relations are informative \citep{newman-griffis-etal-2017-insights}. See Table~\ref{tab:wikiresnli_example} for an example. Their experiment showed that pretrained LMs can predict 29\% of analogous entities in a zero-shot setting and 41\% after training. This indicates that analogical knowledge already exists in pretrained models and can be enhanced by training. 

\paragraph{Generating premises with ``respectively''.} Given four analogical entities $\langle w_1, w_2, w_3, w_4 \rangle$ and the predicate $p$, we form a natural language premise consisting of the analogical information in a respective reading setting of 5(a) after adapting $p$ for phrasing and conjugation. Such a premise is unambiguous and equivalent to 5(b), where the predication is distributed over the two pairs of entities. 5(a) is marked by an explicit respective reading indicator. As an implicit respective reading case, 5(c) has the same meaning as 5(b) but there is no explicit respective operator. In such implicit cases, the predicate $p$ is usually mutually exclusive in that each subject can have only one object. For example, in Sentence 6(a) a person can only die in one place but not two places. 
Non-mutually exclusive predicates are disqualified for an implicit respective reading since they causes ambiguity, as in Sentence 6(b). 


\begin{enumerate}
\setcounter{enumi}{4}
\item
\begin{enumerate}
  \item \small{$w_1$ and $w_3$ $p$ $w_2$ and $w_4$, respectively.}
  \item $w_1$ $p$ $w_2$ and $w_3$ $p$ $w_4$.
  \item $w_1$ and $w_3$ $p$ $w_2$ and $w_4$.
\end{enumerate}
\end{enumerate}

\begin{enumerate}
\setcounter{enumi}{5}
\item
\begin{enumerate}
  \item \small{Emiliano Zapata and Gerhart Münch died in Morelos and Michoacán.}
  \item \small{John and Mary ate a falafel and a tortilla.}
\end{enumerate}
\end{enumerate}



\paragraph{Generating hypotheses.} We subsequently generate hypotheses and pair them with the generated explicit and implicit premises. In Table~\ref{tab:wikiresnli_example}, we show the rules to write entailment or contradiction hypotheses given a premise created from the analogical entities and properties.

\paragraph{Statistics.}
The resulting dataset, which we call WikiResNLI$_\textsc{explicit}$, contains 2,317 premises with different analogical entities, each of which has two entailment hypotheses and six contradiction hypotheses, resulting in 18,536 premise-hypothesis pairs in total. The dataset has 139 different predicates derived from Wikidata properties.
For the development set, we randomly sample 13 predicates from the 126 predicates left and trimmed them if the number of premises for each predicate exceeds 100. We have 1,312 premise-hypothesis pairs for the development set. The rest is used as the training set, with 1,577 premises and 12,616 premise-hypothesis pairs.

\paragraph{Generating premises with implicit ``respectively''.}
We aim to test whether LMs can reason with respective readings and generalize from explicit construction to instances without overt markers. For this purpose, we derive an implicit dataset from WikiResNLI$_\textsc{explicit}$ by simply removing the word ``respectively'' from the premises. We call this dataset WikiResNLI$_\textsc{implicit}$.
In this process, we need to pay special attention to the fact that ambiguity usually occurs in the 1S2O setting when the predicates allow conjunction of objects; given the sentence 6(b), it is ambiguous whether the hypothesis ``John ate a falafel and a tortilla'' is entailed. To form a high-quality test set for WikiResNLI$_\textsc{implicit}$, we first need to exclude the ambiguous contradiction hypotheses. Therefore, two of the authors manually annotate the 139 predicates for whether they allow a single subject predicating conjunction of two objects. In total, 13 predicates are annotated by both authors as unambiguous. Subsequently, we keep only the premises with these predicates from the complete WikiResNLI, and for each predicate, we cap it if the number of premises exceeds 100. Eventually, we are left with 451 premises for the 13 predicates. The 3,608 premise-hypotheses pairs are used as the test set.

\subsection{Naturally-occurring Dataset: NatResNLI}
\label{sec:NatResNLI}

While the synthetic dataset is well-controlled, it does not necessarily cover the natural usage of ``respectively''. To address this, we also collect a dataset of naturally-occuring usages.

\begin{table}[t]
    \centering
    \resizebox{\columnwidth}{!}{%
    \begin{tabular}{lccc}
    \toprule

    \textbf{Human} & Entailment & Neutral & Contradiction \\
    \textbf{Reference} \\
    \midrule
 Entailment & 93.4 & 2.1 & 4.5 \\
 Contradiction & 5.9 & 4.1 & 90 \\

 \bottomrule
     \end{tabular}
}
\caption{NatResNLI human annotated label distribution in percentages for each assigned reference label. Humans mostly agree with the pre-assigned reference labels (demonstrated in Table~\ref{tab:wikiresnli_example}), but not always.}
\label{tab:human-eval}
\end{table}
\paragraph{Collecting premises.} 
As data resources for ``respectively'' in publicly available naturally-occuring data,
we leverage two online dictionaries\footnote{\url{https://sentence.yourdictionary.com/respectively} and \url{https://www.dictionary.com/browse/respectively}} and a writing advice blog,\footnote{\url{https://crosstalk.cell.com/blog/how-to-use-respectively-respectfully}} which provide English examples containing specific words in real-world examples. 
We curate the sentences that included ``respectively'' and further filter some of them to avoid context ambiguity. In total, 76 sentences remain as the premise set.

\paragraph{Generating hypotheses.}
Two of the authors manually write hypotheses based on the fine-grained categorization of Table~\ref{tab:wikiresnli_example} for each collected premise.  Given that the labels are pre-assumed, and to determine whether these inference relations align with humans, we employ crowd workers to verify them. See the annotation details in Appendix \ref{app:annotation detail}.

\paragraph{Statistics.}
The resulting dataset, which we call NatResNLI, consists of 76 premises and 608 hypotheses. 
The average sentence lengths of NatResNLI's premise and hypothesis are 20.1 and 10.1, respectively.  Sentences have 2.32 conjnucts in average, with 4 as the maximum.

\paragraph{Variety.} NatResNLI's sentences have more complicated linguistic constuctions than WikiResNLI, such as relative clauses, e.g., sentence 7(a), implicit coreferences in sentence 7(b), and inverted sentences in sentence 7(c). 

\begin{enumerate}
\setcounter{enumi}{6}
\item
\begin{enumerate}
  \item \small{The annual value of the Hulse endowment is between £800 and £900, of which eight-tenths go to the professor of divinity and one-tenth to the prize and lectureship, respectively.}
  \item In 1910 the export of palm kernels was 6,141 tons, of palm oil 2,160 tons; in 1916 the figures were 22,391 tons and 3,852 tons respectively.
  \item Above this, approached by a stair, are the Lesche and the theatre, occupying respectively the north-east and northwest corner of the precinct.
\end{enumerate}
\end{enumerate}

\paragraph{Inter-annotator Agreement.}

The inter-annotator agreement \citep[Fleiss' kappa;][]{fleiss1971measuring} of the workers for NatResNLI is 0.65, lower than ANLI's (0.67–0.74) and SNLI's (0.70). This can be attributed to that we have five annotators rather than the commonly chosen three annotators, as a larger number of annotators can sometimes lead to more diverse interpretations and disagreements, potentially lowering the inter-annotator agreement.

\paragraph{Verification of pre-assigned labels.}
In Table~\ref{tab:human-eval}, we calculate the average agreement percentage of human annotation with reference labels, showing that humans do not always agree with them. 
Investigating the examples where the majority votes are distinct from the pre-assigned labels, we find nine instances distributed over four premises.
For the sentence in 8(a), humans actually correct the label as the respective reading here does not cause a mutually exclusive effect. For sentence 8(b), humans show more caution towards sentence ambiguity caused by unknown world knowledge of Kilia and Dniester's locations, and hence the neutral label.

\begin{enumerate}
\setcounter{enumi}{7}
\item
\begin{enumerate}
\item \small{\textbf{Premise}: The annual value of imports and exports exceeds seven and nine million sterling respectively. \textbf{Hypothesis}: The annual value of imports and exports exceeds seven million sterling. \textbf{Pre-assigned Label}: contradiction. \textbf{Majority Vote}: entailment}

\item \textbf{Premise}: In that year a Turkish fleet captured the strongholds of Kilia and Akkerman, commanding respectively the mouths of the Danube and Dniester. \textbf{Hypothesis}: In that year a Turkish fleet captured the stronghold of Kilia, commanding the mouths of the Danube and Dniester. \textbf{Pre-asigned Label}: contradiction. \textbf{Majority Vote}: neutral

\end{enumerate}
\end{enumerate}
Considering human annotations as ground truth, we discard the pre-assigned labels and adopt the majority votes as the final labels for NatResNLI.

\begin{table}[t]
    \centering
    \resizebox{\columnwidth}{!}{%
    \begin{tabular}{llccccc}
    \toprule
    
    \textbf{Model} & \textbf{Training data} & entailment & \multicolumn{3}{c}{contradiction} & \textit{overall} \\
     &  & & 1S1O & 1S2O & 2S1O &  \\
    \midrule
 \multirow{2}{*}{RoBERTa} & MNLI & 99.7 & 47.9 & 0.3 & 4.5 & 38.1 \\ 
     & S,M,F,ANLI & 100 & 55.1 & 0.1 & 1 & 39.1  \\ \midrule
\multirow{1}{*}{ALBERT} & S,M,F,ANLI & 99.8 & 31.6 & 3.1 & 3.9 & 34.6 \\ \midrule
\multirow{3}{*}{DeBERTa-v3} & MNLI & 99.4 & 36.1 & 0.6 & 3.9 & 35\\ 
     & M,F,ANLI & 98.8 & 40.2 & 3.9 & 10.8 & 38.4 \\ 
     & M,F,Ling, WANLI & 100 & 77.8 & 36.7 & 59.4 & 68.5\\ \bottomrule
     \end{tabular}
}
\caption{Zero-shot performance on the WikiResNLI$_\textsc{explicit}$ test set. }
\label{tab:zero-shot}
\end{table}

\section{Experiments}
We begin our experiments with the datasets by addressing our first research question:
\begin{theorem}
Can NLI models reason with the coordinate structure in ``respectively'' construction in zero-shot settings?
\end{theorem}

Given the popularity of NLI as a classification task to test LMs' ability of language understanding, many works have proposed new models achieving state-of-the-art results on datasets such as SNLI \citep{bowman-etal-2015-large}, MultiNLI \citep[MNLI;][]{williams-etal-2018-broad} and ANLI \citep{nie-etal-2020-adversarial}. On the GLUE leaderboard,\footnote{\url{https://gluebenchmark.com/leaderboard}} the state-of-the-art models have surpassed 90\% and 95\% accuracy on MNLI and QNLI which are deemed as solved challenges. ANLI has been one of the most challenging tasks in recent years, and the latest models such as DeBERTa-v3-large \citep{He2021deberta, Laurer-etal-2022-less} and PaLM 540B \citep{Huang2022LargeLM} can achieve 64\% and 67.9\%, respectively. While many works use ANLI as a medium to exhibit the models' growing reasoning ability, few of them analyze in depth in which case it fails and at which stage it gets to learn certain linguistic abilities. 

We report the zero-shot performances of three LMs fine-tuned with different combinations of NLI corpora. The models include RoBERTa \citep{liu2019roberta}, ALBERT \citep{Lan2019ALBERTAL} and DeBERTa  with fine-tuning data of MNLI, SNLI, ANLI, FEVER-NLI \citep{nie2019combining}, LingNLI \citep{parrish-etal-2021-putting-linguist} and WANLI \citep{liu-etal-2022-wanli}. 

The experiment results on WikiResNLI$_\textsc{explicit}$ and WikiResNLI$_\textsc{implicit}$ are presented in Table~\ref{tab:zero-shot} and Table~\ref{tab:zero-shot-imp}, respectively.

As can be seen in Table~\ref{tab:zero-shot}, models cannot fully correctly reason with respective readings. The best model, DeBERTa, only achieves 35\% accuracy if fine-tuned with MNLI, and will reach 68.5\% if fine-tuned with almost all NLI training datasets mentioned above. It gains a large increase in the 1S1O setting by 41.7\%. However, the accuracy on 1S2O is still at a chance level, and the 2S1O setting performance is only approaching around 60\%, leaving room for improvement. 
\begin{table}[t]
    \centering
    \resizebox{\columnwidth}{!}{%
    \begin{tabular}{llccccc}
    \toprule
    
    \textbf{Model} & \textbf{Training data} & entailment & \multicolumn{3}{c}{contradiction} & \textit{overall} \\
     &  & & 1S1O & 1S2O & 2S1O &  \\
    \midrule
 \multirow{2}{*}{RoBERTa} & MNLI & 97.1 & 26.4 & 0.4 & 8.6 & 33.1 \\ 
     & S,M,F,ANLI & 99.9 & 23.5 & 0.3 & 3.4 & 31.8  \\ \midrule
\multirow{1}{*}{ALBERT} & S,M,F,ANLI & 100 & 14 & 0.1 & 0.8 & 28.7 \\ \midrule
\multirow{3}{*}{DeBERTa-v3} & MNLI & 99.3 & 25.5 & 1.4 & 5.2 & 32.9\\ 
     & M,F,ANLI & 96.9 & 26.6 & 5.7 & 16.9 & 36.5 \\ 
     & M,F,Ling, WANLI & 100 & 59.3 & 2.4 & 13.2 & 43.7\\ \bottomrule
     \end{tabular}
}
\caption{Zero-shot performance on the WikiResNLI$_\textsc{implicit}$ test set. }
\label{tab:zero-shot-imp}
\end{table}

The performance on WikiResNLI$_\textsc{implicit}$ is even worse, as indicated in Table~\ref{tab:zero-shot-imp}. Similarly, DeBERTa is again the best performance model on the dataset, with an accuracy of 43.7\% if fine-tuned with all NLI corpora. The accuracy is just 10\% above the chance level, and it completely fails in the 1S2O and 2S1O settings.

Results on both datasets show that when training with more data, models improve on respective readings. However, the question of what leads to improvement remains. We examine how many times explicit respective readings appear in the training and testing datasets of MNLI, SNLI Fever-NLI and ANLI. We find that the adverb ``respectively'' occurs 177 and 12 times in the MNLI training and dev sets, 15 and 0 times in the SNLI training and test sets, 1,064 and 64 times in the Fever-NLI training and test sets, and 216 and 5 times in the combined ANLI training and dev sets. We randomly sampled a subset of each dataset and manually check whether they tackle reasoning over coordination structure. We find that in most cases, ``respectively'' works simply as a context word and has little to do with the actual inference relations. Thus it is still not clear whether it is simply the exposure to the explicit cues (the word ``respectively'') or some instances with implicit coordinate structures that result in the performance improvement. We thus ask the following three research questions and experiment with few-shot learning.

\begin{figure}[!t]
    \centering
    \includegraphics[width=\columnwidth]{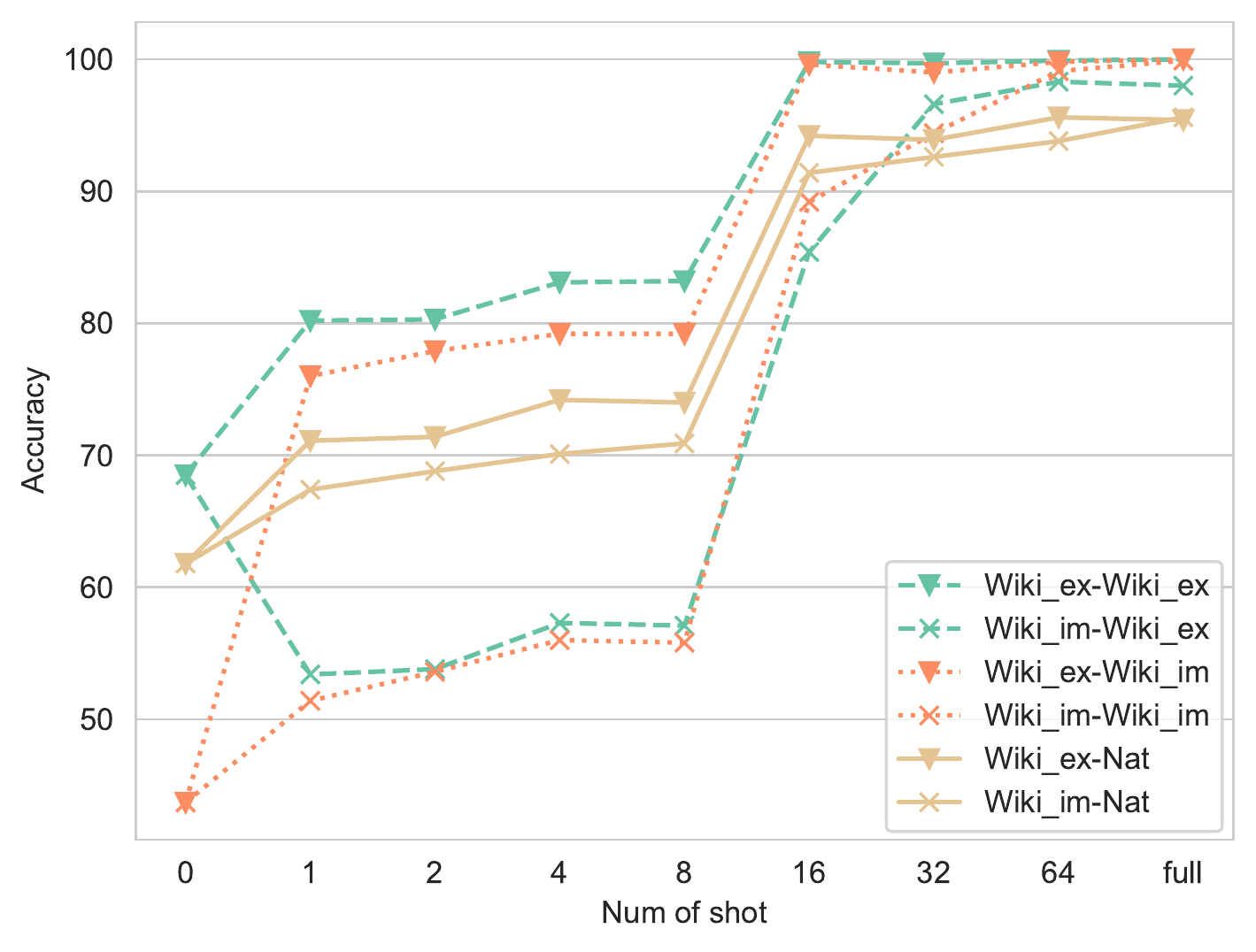}
    \caption{Overall performance of DeBERTa on WikiResNLI$_\textsc{explicit}$, WikiResNLI$_\textsc{implicit}$ and NatResNLI from zero-shot to fully supervised. \textit{Wiki\_ex-Wiki\_ex} refers to training with WikiResNLI$_\textsc{explicit}$ instances and evaluating on WikiResNLI$_\textsc{explicit}$ test set. Similarly, \textit{Wiki\_im-Nat} refers to training with WikiResNLI$_\textsc{implicit}$ and testing on NatResNLI.  }
    \label{fig:oneplot-overall-accuracy}
\end{figure}
\begin{table}[h]
    \centering
    \resizebox{\columnwidth}{!}{%
    \begin{tabular}{lccccccccc}
    \toprule
    \textbf{\# Shots} & 1 & 2 & 4& 8 & 16 & 32 & 64 & Full \\
    \textbf{Type} \\
    \midrule
All & 8 & 16 & 32 & 64 & 128 & 256 & 512 & 12,616 \\
Basic & 4& 8 & 16 & 32 & 64 & 128 & 256 & 6,308 \\
 \bottomrule
     \end{tabular}
}
\caption{Number of training instances for each number of shots. A ``shot'' contains multiple training instances since we always take a premise along with all of its generated hypotheses---8 in the general case and 4 in the basic case.}
\label{tab:nums_of_training}
\end{table}


\begin{theorem}
Can LMs Generalize from Explicit to Implicit Respective Readings?
\end{theorem}

Instances of WikiResNLI have the coordinate structures of an equal number of conjuncts, and linguists have argued that such semantic relations are reflected in the syntactic relations \citep{Goodall2009ParallelSI,Moltmann1992CoordinationAC}. 
It is essentially semantic but also relies on pragmatically available information of the truth conditions.
Respective readings in fact also commonly omit explicit lexical indicators but remain available and preferred as 2(a) \citep{Gawron2004TheSO}. We are therefore interested in whether LMs can learn the semantic-pragmatic meaning of respective reading sentences rather than only making use of lexical and syntactic cues.

We fine-tune the DeBERTa model previously fine-tuned with M, F, Ling and WANLI with different numbers of WikiResNLI$_\textsc{explicit}$ examples without a dev set, since we do not want to bias the model towards our datasets hence hurting performance on the other NLI tasks. 
\begin{figure}[!t]
    \centering
    \includegraphics[width=\columnwidth]{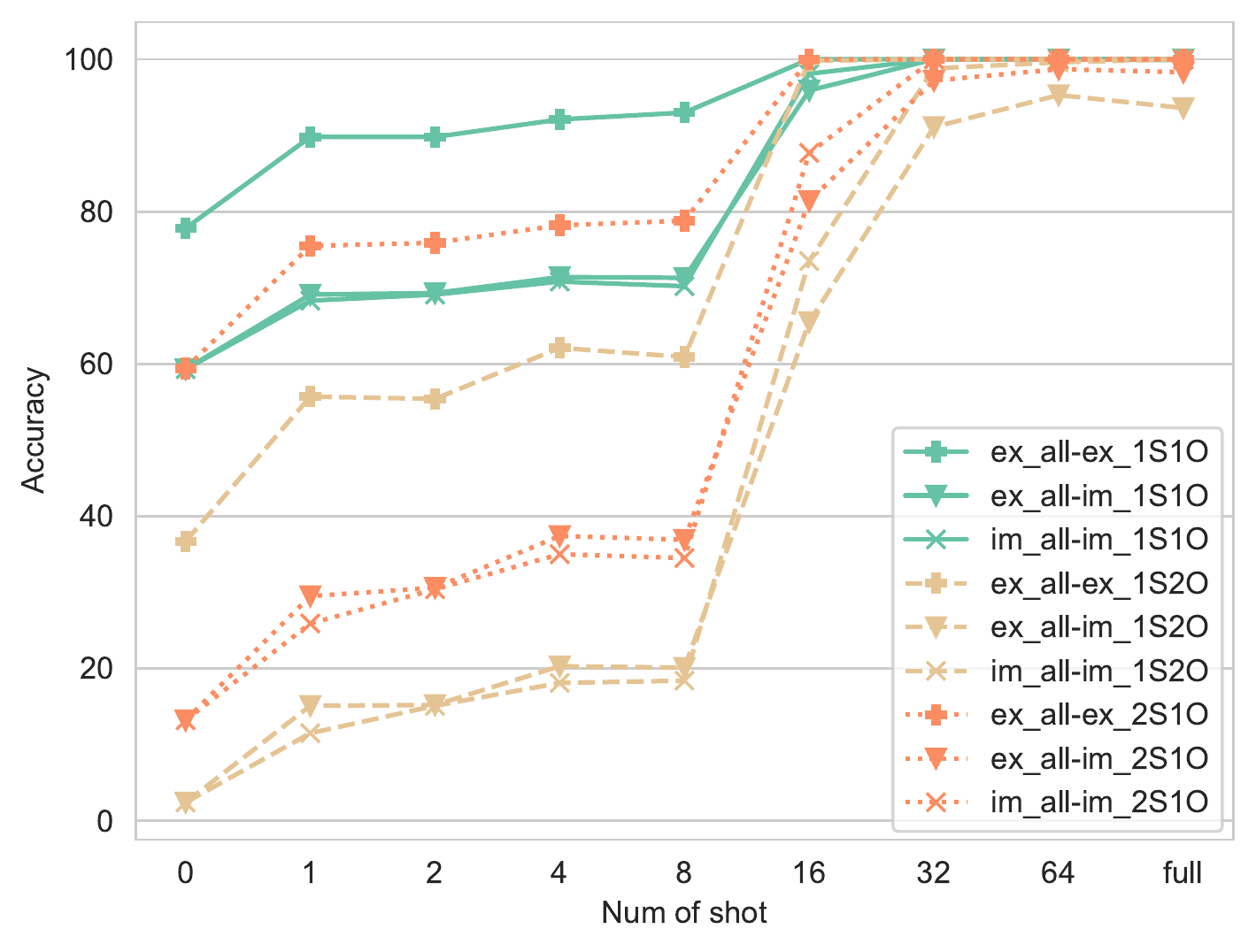}
    \caption{DeBERTa's Performances on WikiResNLI$_\textsc{implicit}$ after fine-tuning on WikiResNLI$_\textsc{explicit}$ or WikiResNLI$_\textsc{implicit}$. The result is broken down by contradiction fine-grained set.}
    \label{fig:oneplot-ex2im}
\end{figure}
We fine-tune the model with  WikiResNLI$_\textsc{explicit}$
and  WikiResNLI$_\textsc{implicit}$ separately and report the overall accuracy on both dataset in Figure \ref{fig:oneplot-overall-accuracy}. Training with WikiResNLI$_\textsc{explicit}$ contributes to a steady performance increase on both WikiResNLI$_\textsc{explicit}$ and WikiResNLI$_\textsc{implicit}$. Especially, 1-shot learning enhances the performance clearly, with a 10\% increase for in-domain evaluation, and a remarkable 30\% increase for explicit to implicit generalization. The improvements are small from 1-shot to 8-shot. Only at 16-shot, both WikiResNLI$_\textsc{explicit}$ in-domain learning and transferring to WikiResNLI$_\textsc{implicit}$ reach 100\% accuracy. This shows the possibility to learn respective readings, despite the need to see relevant instances 128 times (see Table~\ref{tab:nums_of_training}). 

Interestingly, in-domain few-shot learning of WikiResNIL$_\textsc{implicit}$ witnesses a relatively cold start. The accuracy does not increase above 60\% until 16 shots. Generalization from implicit respective reading to explicit reading is surprisingly not reaching 100\% accuracy even after full supervision. We are keen to investigate what types of instances are difficult to learn for explicit to implicit respective reading generalization. In Figure \ref{fig:oneplot-ex2im}, 
we break down WikiResNLI$_\textsc{implicit}$ with contradiction labels by categories (1S1O, 1S2O and 2S1O) and plot the accuracy against number of shot.

As can be seen, the performance on explicit readings is always better than on implicit readings across all three contradiction types. Among them, 1S2O and 2S1O instances are the most difficult. Their accuracies are below 40\% and 20\%, respectively before 16 shots. And only until 32 shots do both types reach above 95\% accuracy. Unlike in-domain learning, 1S2O never gets perfectly solved.


\begin{figure}[!t]
    \centering
    \includegraphics[width=\columnwidth]{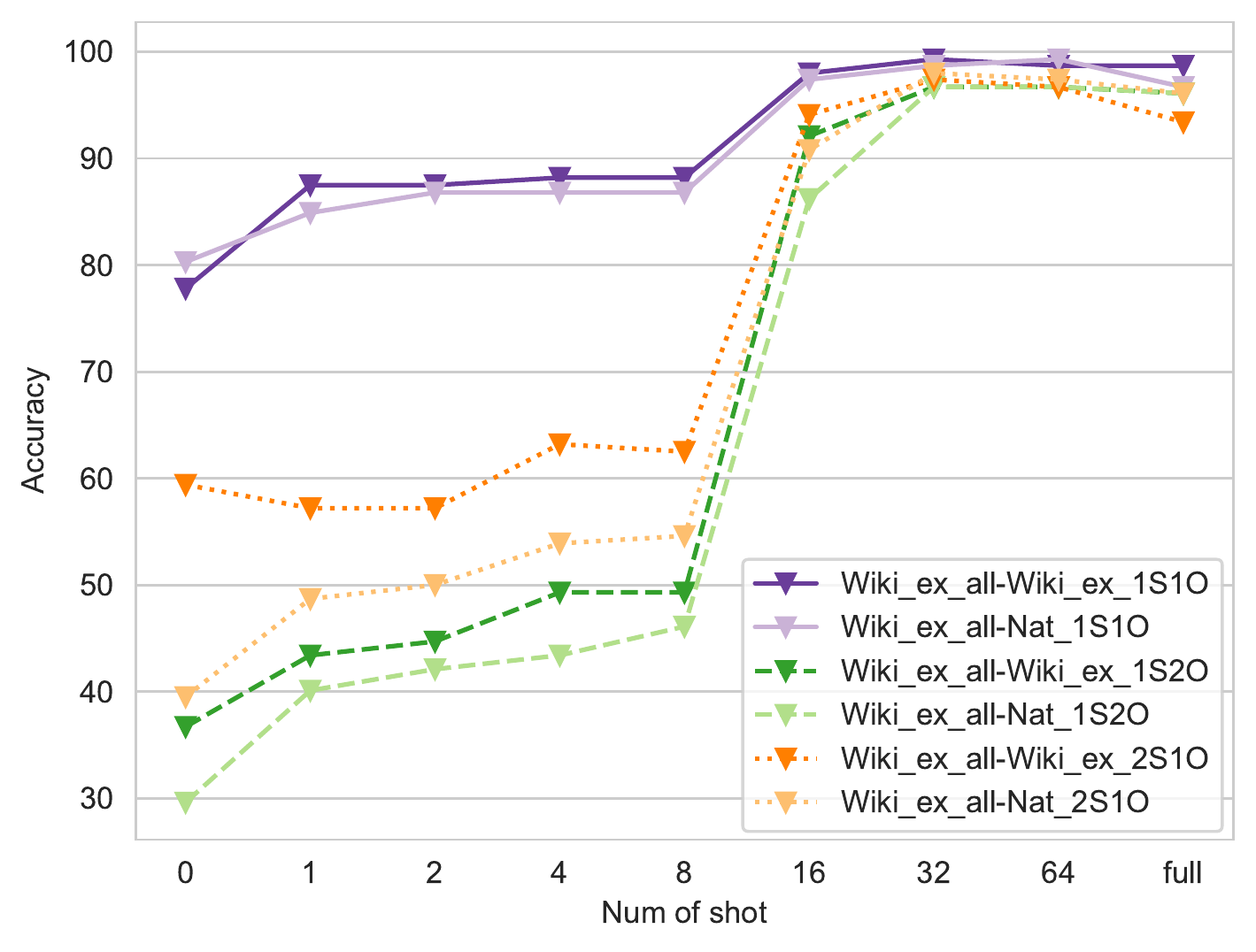}
    \caption{Performance of DeBERTa on NatResNLI after being fine-tuned on WikiResNLI$_\textsc{explicit}$. To facilitate comparison, we mark performances on WikiResNLI$_\textsc{explicit}$ in darker colours. }
    \label{fig:oneplot-wiki2nat}
\end{figure}

\begin{theorem}
Can LMs Generalize from Synthetic to Natural Respective Readings?
\end{theorem}

WikiResNLI is a synthetic dataset, and it remains unclear whether models can reason with respective readings in realistic settings if we generate enough synthetic data and feed it to models. With NatResNLI, we are able to investigate LM's respective reading reasoning generalizability from synthetic to natural data and its alignment with humans. 

We evaluate the models fine-tuned with WikiResNLI$_\textsc{explicit}$ on NatResNLI and plot the performance in Figure \ref{fig:oneplot-wiki2nat}. We can observe that scores on NatResNLI are almost always lower than on WikiResNLI due to domain drift. Particularly, 1S2O and 2S1O are 10\% and 20\% lower in zero-shot settings. 1S2O manage to reach on-par performance with WikiResNLI after 16 shots, while 2S1O after 32 shots. 

Interestingly, the models are able to surpass 95\% after 32 shots, while pre-assigned labels only have 90\% match (see Table~\ref{tab:human-eval}). Although we are comparing a rule-based method with 32-shot (256 examples) training, we can conclude that models are able to align with humans for respective reading reasoning. In addition, we notice that for 1S2O and 2S1O generalization, the complex linguistic structures discussed in Section \ref{sec:NatResNLI} do have a high impact in the low-number few-shot learning, but the difficulty diminished as more training data are used. 

\begin{figure}[!t]
    \centering
    \includegraphics[width=\columnwidth]{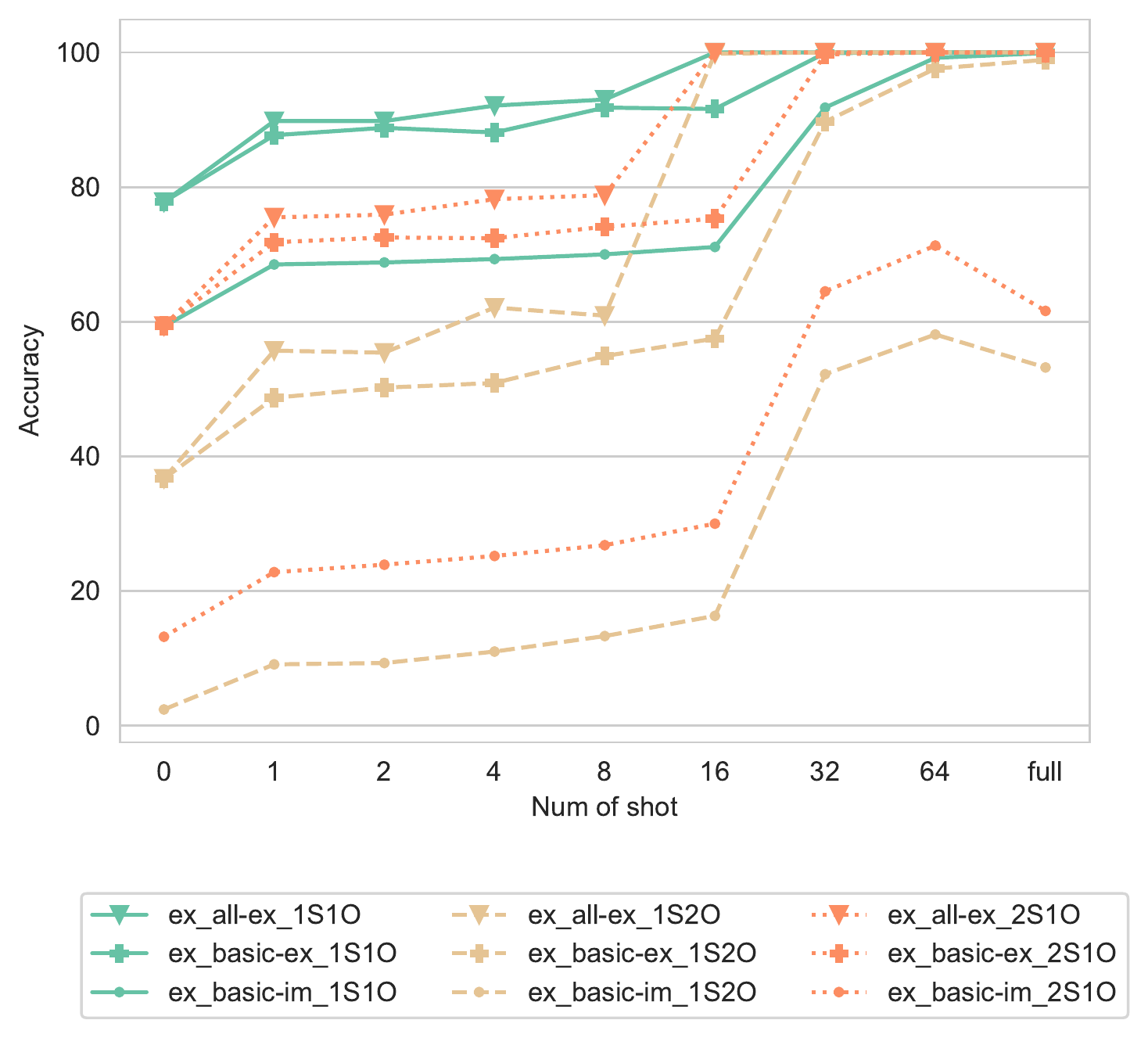}
    \caption{Performance of DeBERTa on WikiResNLI$_\textsc{explicit}$ and WikiResNLI$_\textsc{implicit}$ after being fine-tuned only with the basic types (entailment and 1S1O contradiction) of WikiResNLI$_\textsc{explicit}$.}
    \label{fig:oneplot-base2all}
\end{figure}

\begin{theorem}
What Cues do LMs Rely on?
\end{theorem}

So far we have discussed LMs' ability to generalize on the syntactic-semantic level, from explicit to implicit and from synthetic to natural in respective readings. But it is yet to be determined whether the model is simply adopting the lexical-syntactic heuristics for prediction and whether it leverages common sense and world knowledge. If models can reason over basic hypothesis structures (1S1O entailment and 1S1O contradiction), it would be expected they are aware that the one-to-one relation correspondences should exclude 1S2O and 2S1O propositions due to common sense and world knowledge. Although there are cases such as 8(a) where one object entity includes the other in NatResNLI, all cases of the WikiResNLI test set disallow the situation due to the mutually exclusive properties. 

Therefore, we fine-tuned the DeBERTa models with only WikiResNLI$_\textsc{explicit}$ instances of basic structures and evaluated their performances on both WikiResNLI$_\textsc{explicit}$ and WikiResNLI$_\textsc{implicit}$ 1S2O and 2S1O. The results can be seen in Figure \ref{fig:oneplot-base2all}. We can observe that the generalization from basic structures to unseen structures is indeed difficult: while training with all structures and evaluating will all structures achieve perfect scores on 1S2O and 2S1O of WikiResNLI$_\textsc{explicit}$ at 16 shots, training with basic structures are only 58\% and 75\% accuracies. It is worth noting that all fine-tuning instances have either entailment or contradiction labels, and therefore a random-guessing baseline would be 50\% instead of 33.3\%. 

\begin{figure}[!t]
    \centering
    \includegraphics[width=\columnwidth]{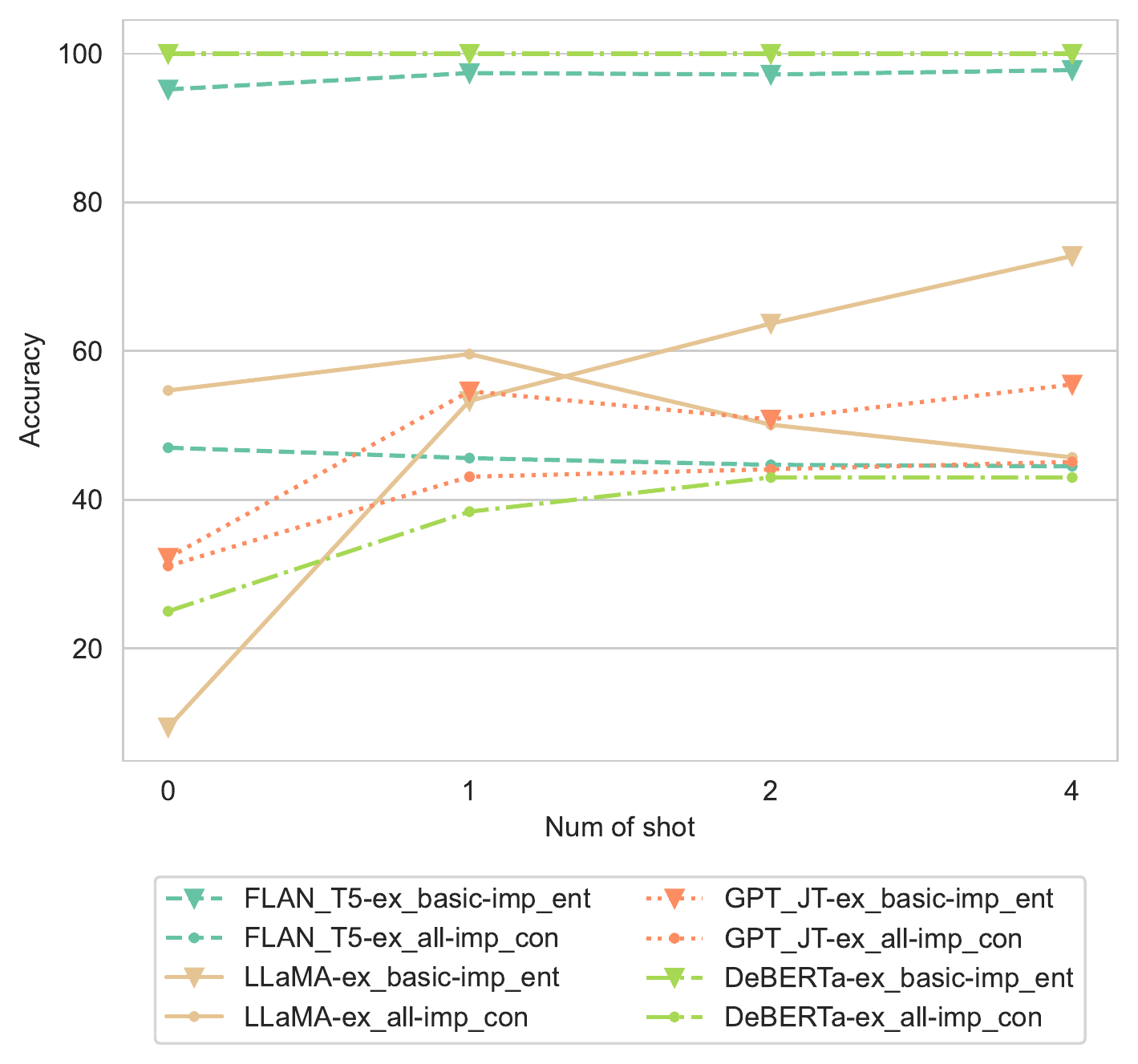}
    \caption{LLaMA, FLAN-T5, GPT-JT and DeBERTa's performances on WikiResNLI$_\textsc{implicit}$ after in-context learning of WikiResNLI$_\textsc{explicit}$. The last suffix \textit{ent} of a legend means the performance on entailment pairs and \textit{con} on contradiction pairs.}
    \label{fig:oneplot-llm_ex2im}
\end{figure}

The generalization performances from explicit respective readings with basic structures to implicit 1S2O and 2S1O are more disappointing. At 16 shots, the accuracies are only 18\% and 30\%, respectively, well below the chance level. Even full supervision can only achieve around 60\% accuracy for both structures. The results indicate that the models do not effectively learn the abstract respective reading relations due to not understanding the commonsense and world knowledge. 

We look into the intersection errors of 32-shot, 64-shot and fully-supervised models which are fine-tuned on WikiResNLI$_\textsc{explicit}$ and are evaluated on WikiResNIL$_\textsc{implicit}$. There are 358 1S2O and 248 2S1O instances that are consistently mistaken by the models. The top-5 frequent properties are: twinned administrative bodies, took place, are capitals of, buried in, and family names. Knowledge about relative location  9(a) and knowledge about humans 9(b) thus seem to play an important role in reasoning with implicit respective readings.

\begin{figure}[!t]
    \centering
    \includegraphics[width=\columnwidth]{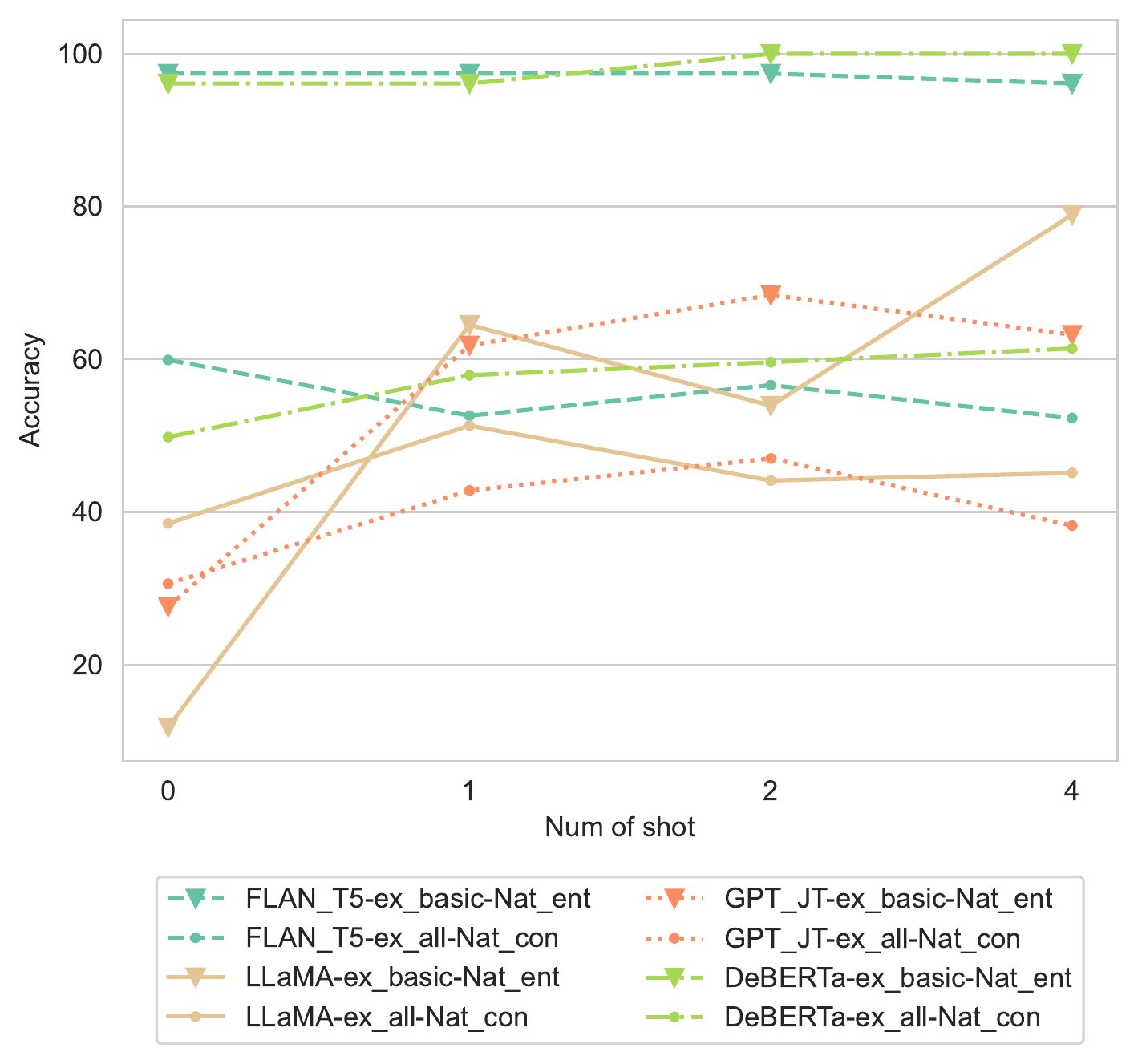}
    \caption{LLaMA, FLAN-T5, GPT-JT and DeBERTa's performances on NatResNLI after in-context learning of WikiResNLI$_\textsc{explicit}$.}
    \label{fig:oneplot-llm_wiki2nat}
\end{figure}

\begin{enumerate}
\setcounter{enumi}{8}
\item
\begin{enumerate}
\item \small{\textbf{Premise}: Battle of Tours and Battle of Verdun took place in Poitiers and Verdun. \textbf{Hypothesis}: Battle of Tours took place in Poitiers and Verdun. \textbf{WikiResNLI Label}: contradiction. \textbf{Prediction}: entailment}

\item \textbf{Premise}: Theresa of León and Maria Solé Cuñat died in Galicia and Catalonia.  \textbf{Hypothesis}: Theresa of León and Maria Solé Cuñat died in Galicia. \textbf{WikiResNLI Label}: contradiction. \textbf{Prediction}: entailment

\end{enumerate}
\end{enumerate}

\paragraph{Impact on other NLI tasks.}
We evaluate all models fine-tuned with WikiResNLI above on other NLI tasks, i.e, MNLI-m and ANLI-R3, to check whether fine-tuning on such a label-imbalanced dataset hurts performance. Interestingly, full supervision with WikiResNIL$_\textsc{implicit}$ of basic structures results in new state-of-the-art performance for DeBERTa. On MNLI-m, the score improves from 90.8\% to 91.4\%; and on ANLI-R3, the performance raises from 63.6\% to 64.1\%.

\paragraph{Experiments on LLaMA, FLAN-T5 and GPT-JT}
Significant advancements in large generative LMs have been achieved in the realm of general natural language understanding. These improvements can be attributed to enhanced training strategies, such as incorporating code and human instructions into pretraining/fine-tuning data and RLHF \citep{christiano2017deep,openai2023gpt4}. 
We assess the zero-shot and in-context learning abilities of three open-source generative models, that is, LLaMA-7B \citep{Touvron2023LLaMAOA}, FLAN-T5-XL \citep{Chung2022ScalingIL} and GPT-JT-6B \citep{gpt-j,gpt-jt}. In this study, our focus is on two representative scenarios, namely generalizing from explicit to implicit readings and generalizing from synthetic to natural readings. We adopt the template \textit{\{premise\} Question: Does this imply that \{hypothesis\}?} as it attains top-tier results for NLI tasks \citep{webson-pavlick-2022-prompt}.

Figure \ref{fig:oneplot-llm_ex2im} illustrates the explicit to implicit generalization results. Notably, FLAN-T5 achieved a near-perfect score on zero-shot entailment pairs, comparable to the fine-tuned DeBERTa. However, GPT-JT, despite being instruction-tuned on NLI datasets, performed at a mere chance level on entailment pairs, while LLaMA scored below 10\% accuracy. In terms of contradiction instances, all three models scored below 60\% accuracy, with in-context learning offering limited improvement at the 4-shot level. Specifically, FLAN-T5's performance decreased after in-context learning.

For the generalization from WikiResNLI to NatResNLI, in Figure \ref{fig:oneplot-llm_wiki2nat}, we observed similar trends as in the previous experiments. FLAN-T5 outperformed the other models on entailment instances, and LLaMA demonstrated significant improvement within a few shots. However, for contradiction pairs, all models experienced only a modest increase in accuracy from 1 to 4 shots, with the highest accuracy remaining below 60\%. 

To conclude, while large generative models have made significant strides in natural language understanding, they still face substantial challenges in reasoning with respective readings, highlighting the need for further research and development in the long tail of linguistic constructions.

\section{Related Work}
Logical relations between two sentences are a core aspect of language understanding \citep{FregeBegriffsschriftED,heijenoort1967logic, blackburn2006handbook}. To facilitate large-scale model evaluation, NLP researchers have developed manually labelled NLI corpora, typically for 2/3-way classification \citep{dagan2013recognizing, bowman-etal-2015-large, williams-etal-2018-broad}. In recent years, researchers start to analyze the characteristics of these datasets, such as annotation artefacts \citep{gururangan-etal-2018-annotation}, syntactic heuristics \citep{mccoy-etal-2019-right} and adversarial collection process \citep{williams-etal-2022-anlizing}.

In computational linguistics, distributive predication has been analyzed through means of distributivity operators \citep{massey1976tom,link1983logical,roberts1987modal,lasersohn1998generalized}. And linguists have been working on extending first-order logical forms to include distributive and collective readings \citep{martin-1981-formal, alshawi-van-eijck-1989-logical}. \citet{scha-stallard-1988-multi} present a recursive translation rule scheme to account for multi-level plurals. \citet{aone-1991-resolution} proposed a reasoner consisting of domain-dependent constraints and domain-independent axioms for collective and distributive ambiguity. \citet{shaw-mckeown-2000-generating} described a simplified quantifier system to minimize distributive and collective ambiguities. 

Respective readings have not yet been studied in modern NLP. Relevant works include plural understanding, which has been studied as a coreference resolution task  \citep{jain-etal-2004-anaphora,zhou-choi-2018-exist, yu-etal-2020-free}.  \citet{manshadi-etal-2011-corpus} proposed quantifier scope annotation 
in which plurals are annotated with distributive and collective readings at the constraint level. 
\citet{yu-etal-2020-word} show that LMs are better at reflexive anaphora tasks with distributive than collective constructions.

\section{Conclusions}
The ``respectively'' construction is simple yet entails multiple levels of reasoning skills, including syntactic-semantic and commonsense-world knowledge. It is crucial that when an out-of-the-box model cannot reason over it, it should be able to learn with as few examples as possible. We proposed two datasets, WikiResNLI (a controlled synthetic dataset) and NatResNLI (a naturally occurring dataset) to probe their ability to do so 
in zero-shot and few-shot settings. We find that 
explicit reasoning is easier to learn than implicit reasoning, and LMs fail to generalize when common sense inference is needed. 
We confirm that diverse and complex training data are necessary to achieve human-level performance.

\section{Limitation}
Linguistic studies have shown that respective readings are not necessary to have two coordinate structures in the same sentence \citep{dalrymple1995constraints}. Both WikiResNLI and NatResNLI have only one sentence in the premise and do not exhaust all possible and complicated realizations of respective readings. However, we are able to discuss and investigate LMs' generalizability with ``respectively'' with three constructions, i.e., 1S1O, 1S2O and 2S1O. 

Our experiments are English-specific and are limited to LMs that can be run with an academic budget. However, our conclusions about the generalizability towards respective readings should be viewed as language-agnostic given there are linguistic constructions under-discussed in many other languages and it is worth researchers' attention to study them.

\section{Acknowledgments}
We would like to thank the members of the CoAStaL NLP group and the anonymous reviewers for their helpful suggestions.

\bibliography{anthology,references}
\bibliographystyle{acl_natbib}
\clearpage

\appendix
\section*{Appendices}
\label{sec:appendix}

\section{Annotation Details}
\label{app:annotation detail}
We employ Amazon Mechanical Turk workers. A qualified worker is one who has completed more than 10,000 HITs and has an approval rate greater than 99\%. We set the location to the United States as there was no option to choose language proficiency. They are shown only three examples with entailment, neutral and contradiction labels before annotation. For each premise-hypothesis pair, five workers were asked to annotate the entailment relation (entailment, neutral or contradiction) following the guidelines of \citet{nie-etal-2020-adversarial}. The worker gains a reward of 12 cents. Based on the workers' feedback, our hourly rate ranges between 16 to 27 US dollars, which is above the federal or Californian hourly wage. In total, 170 annotators participated in the step of label annotation of the hypotheses written by the authors. The number of HITs (annotation) per worker ranges from 5 to 200 based on their wishes. We assure to have 5 annotations per each premise-hypothesis pair. 

\end{document}